\documentclass[preprint,12pt]{elsarticle}

\usepackage{amssymb}
\usepackage{amsmath}
\usepackage{bm}
\usepackage[utf8]{inputenc}
\usepackage[justification=centering]{caption}
\usepackage{epsfig}
\usepackage{float}
\usepackage{tabularx}
\usepackage{subfig}
\usepackage{graphicx}
\usepackage{diagbox}[2011/11/22]

\journal{Neurocomputing}

\begin{document}

\begin{frontmatter}



\title{Co-regularized Multi-view Sparse Reconstruction Embedding for Dimension Reduction}


\cortext[cor1]{Corresponding author: Xianping Fu}
\author[1]{Huibing Wang}
\author[1]{Jinjia Peng}
\author[1]{Xianping Fu\corref{cor1}}

\address[1]{Information Science and Technology College, Department of Computer Science and Technology, Dalian Maritime University, Dalian, Liaoning, 116026, China.}

\begin{abstract}
With the development of information technology, we have witnessed an age of data explosion which produces a large variety of data filled with redundant information. Because dimension reduction is an essential tool which embeds high-dimensional data into a lower-dimensional subspace to avoid redundant information, it has attracted interests from researchers all over the world. However, facing with features from multiple views, it's difficult for most dimension reduction methods to fully comprehended multi-view features and integrate compatible and complementary information from these features to construct low-dimensional subspace directly. Furthermore, most multi-view dimension reduction methods cannot handle features from nonlinear spaces with high dimensions. Therefore, how to construct a multi-view dimension reduction methods which can deal with multi-view features from high-dimensional nonlinear space is of vital importance but challenging. In order to address this problem, we proposed a novel method named Co-regularized Multi-view Sparse Reconstruction Embedding (CMSRE) in this paper. By exploiting correlations of sparse reconstruction from multiple views, CMSRE is able to learn local sparse structures of nonlinear manifolds from multiple views and constructs significative low-dimensional representations for them. Due to the proposed co-regularized scheme, correlations of sparse reconstructions from multiple views are preserved by CMSRE as much as possible. Furthermore, sparse representation produces more meaningful correlations between features from each single view, which helps CMSRE to gain better performances. Various evaluations based on the applications of document classification, face recognition  and image retrieval can demonstrate the effectiveness of the proposed approach on multi-view dimension reduction. 

\end{abstract}


\end{frontmatter}


\section{Introduction}
In recent years, with the development of information technology, researchers have proposed a plenty of algorithms or equipments to describe samples from various perspectives \cite{li2012multiview,hoi2013online}. For one single sample, multi-view features can reflect different properties of one same sample from different viewpoints. Take image processing as examples, one image can be represented by  multiple feature descriptors, such as Local Binary Patterns (LBP)~\cite{huang2011local}, Locality-constrained Linear Coding (LLC)~\cite{wang2010locality}, Histograms of Oriented Gradients (HOG) \cite{dalal2005histograms}, etc~\cite{bi2016multi,yang2016fast}. Even though features extracted from different descriptors show multiple properties, they are all descriptions of one same image, which contain some common interior relations between each other. Specifically, because more information exists in multiple views rather than in one single view, how to fully integrate compatible and complementary information from multiple views, which can make better decisions for whole system is of vital importance but challenging. 

For most research fields, such as image classification~\cite{hoi2006batch,wu2018deep}, image retrieval~\cite{tao2006direct,passalis2017learning,spyromitros2014comprehensive,gao2014soml,wang2017effective}, image understanding~\cite{tao2007general,wu2018andwhere}, most features which are utilized in these fields always locate in high-dimensional spaces. Therefore, it will consume much time and computation space to cope with multi-view features with high dimensions. It's essential for researchers to construct meaningful dimension reduction (DR) methods which can find optimal subspaces to maintain some properties of original features into subspaces with lower dimensions. To address this problem, there are many effective DR methods have been proposed, which can be categorized into two divisions. The first category contains various linear DR methods, such as Principal Component Analysis (PCA)~\cite{wold1987principal}, Linear Discriminant Analysis (LDA)~\cite{zollanvari2011analytic}, Locality Preserving Projections (LPP)~\cite{he2004locality}, etc~\cite{he2005neighborhood,qiao2010sparsity,laohakiat2017clustering}. PCA~\cite{wold1987principal} and LDA are two most famous ones which can exploit global properties of data and maintain them into low-dimensional subspace. However, both of them care more about global Euclidean structure while neglecting local information between adjacent samples. To address this problem, Locality Preserving Projections (LPP)~\cite{he2004locality} and Neighborhood Preserving Embedding (NPE)~\cite{he2005neighborhood} are proposed to maintain adjacent structure using different reconstruction tricks. Meanwhile, some sparse subspace learning methods~\cite{qiao2010sparsity} have also arose much attentions. Sparsity Preserving Projections (SPP)~\cite{qiao2010sparsity} is a well-known methods which exploits sparse correlations \cite{zhou2010exclusive} between samples and preserves them into a sparse subspace. However, all DR methods mentioned above are linear DR ones which cannot deal with nonlinear datasets well. Locally Linear Embedding (LLE)~\cite{roweis2000nonlinear} and Laplacian Eigenmaps (LE)~\cite{belkin2003laplacian}  are two popular nonlinear dimension reduction techniques, which aim to obtain the representations with lower dimensions directly by preserving certain correlations between samples from nonlinear manifolds. Even though there are many DR methods which can deal with various circumstances, most of them can only utilize features from one single view but multiple views. They failed to integrate compatible and complementary information from multi-view features and construct optimal subspaces with lower dimensions, which is a hot research topic during the last decades.

The past decade has also witnessed the blossom of multi-view learning in many practical areas, such as clustering~\cite{wang2018multiview,kumar2011co,xia2010multiview,xu2015multi,wang2015robust,wang2016iterative}, metric learning~\cite{yu2016deep}, etc~\cite{wu2019few,kumar2011co,hong2015multi,wu20193}. Xia et al.~\cite{xia2010multiview} proposed a multi-view spectral embedding (MSE) which fully encodes different features in different ways and finds a low-dimensional embedding wherein the distribution of each view is sufficiently smooth. Kumar et al.~\cite{kumar2011co} develops a co-regularized scheme to combine features from multiple view together and obtain good results for spectral clustering. Meanwhile, they proposed an iterative procedure to deal with the co-regularized optimization problem. Xu et al.~\cite{xu2015multi} proposed a novel method called Multi-view Self-Packed Learning (MSPL) which adopts a novel probabilistic smoothed weighting scheme for clustering. Multi-view discriminant analysis~\cite{kan2016multi} is an extension of traditional LDA~\cite{zollanvari2011analytic}, which projects features from multiple views to one discriminative common subspace. However, as we all known, there is few multi-view DR method which directs at nonlinear datasets to find their low-dimensional representations. 

In this paper, we proposed a multi-view nonlinear DR method named Co-regularized Multi-view Sparse Reconstruction Embedding (CMSRE) to deal with the problem of multi-view nonlinear dimension reduction. CMSRE first exploits the correlations of sparse reconstruction between features from each single view and obtains the initial low-dimensional representations. Then, it utilizes a co-regularized scheme to update the low-dimensional representations of all views. It seems to be worth mentioning that CMSRE considers the correlations of sparse reconstructions from all views while updating the low-dimensional representations of each single view. Therefore, the low-dimensional representation of each single view can fully utilize the information from the other views, which can improve the representational abilities of CMSRE. Finally, an iterative solving procedure is constructed to obtain optimal low-dimensional representations for CMSRE. And we summarized the contributions of this paper as follows:

\begin{itemize}
	\item We have carefully surveyed the filed of multi-view DR method and proposed a method named Co-regularized Multi-view Sparse Reconstruction Embedding (CMSRE) to fill the blank of this field.
	\item CMSRE fully considers correlations of sparse reconstruction from all views and maintains these correlations into low-dimensional subspaces for all views as much as possible.   
	\item In order to update low-dimensional representations of features, we introduced a co-regularized scheme to integrate compatible and complementary information from all views. 
\end{itemize}

The rest of paper is organized as follows: in Section \uppercase\expandafter{\romannumeral2}, we introduced some basic elements of multi-view learning and some typical related works. Section \uppercase\expandafter{\romannumeral3} described the construction procedure of our proposed CMSRE to deal with the problem of multi-view nonlinear DR problem. In Section \uppercase\expandafter{\romannumeral4}, we have carried out various experiments on applications of document classification, face recognition \cite{hoi2013fans} and image retrieval to demonstrate the effectiveness of our proposed CMSRE. Then, we added an experiment to show the convergence property of our proposed method. Finally, Section \uppercase\expandafter{\romannumeral5} made a conclusion of this paper.

\section{Related works}

In this section, we introduced some element backgrounds of multi-view learning in detail to help readers to improve the readability of this paper. Then, a multi-view embedding method called Multiview Spectral Embedding (MSE)~\cite{xia2010multiview} is illustrated, which has attracted widely attentions. 

\subsection{Background} 
As mentioned above, multi-view dimension reduction is a hot topic and has been started on in the past decade. Therefore, it's essential to introduce the background of this field. For multi-view learning methods, multi-view features with $n$ samples having $m$ views can be represented as $X=\{X^{(v)}\in \Re^{m_{v}\times n}\}^m_{v=1}$, where $X^{(v)} = [x^{(v)}_1, x^{(v)}_2,\cdot \cdot \cdot, x^{(v)}_n]\in \Re^{m_{v}\times n}$ is a feature matrix which consists of features from the $v$th view. $m_v$ is the dimension of features from $v$th view. However, features from multiple views always locate in high-dimensional spaces with different dimensions, which cannot be utilized directly due to the high time and space complexities. Meanwhile, multi-view features contains integrate compatible and complementary information, which can improve the performances of traditional single-view approaches greatly. Therefore, the goal of our proposed methods is to construct an excellent architecture to consider features from multiple views simultaneously. As described, CMSRE aims to obtain low-dimensional representations as $Y=\{Y^{(v)}\in \Re^{d\times n}\}^m_{v=1}$, where $Y^{(v)} = [y^{(v)}_1, y^{(v)}_2,\cdot \cdot \cdot, y^{(v)}_n]\in \Re^{d\times n}$ is the low-dimensional representations for the $v$th view and $d$ is dimension of common subspace constructed by CMSRE.

\subsection{Multiview Spectral Embedding (MSE)}

Multiview Spectral Embedding (MSE) is a well-known multi-view DR method which has attracted widely attentions from researchers. MSE finds a low-dimensional embedding wherein the distribution of each view is sufficiently smooth, and MSE explores the complementary property of different views. It first builds a patch for each features on a view and performs the procedure of the single-view optimization to get the optimal low-dimensional embedding for each view. Finally, MSE proposed a method called global coordinate alignment to unify all low-dimensional embeddings from different patches as a whole one. Alternating optimization is utilized to obtain the solution for MSE. In order to finish its aim, MSE constructs the objective function as follows:

\begin{equation}
\label{eq1}
\begin{array}{l}
\mathop {\arg \min }\limits_{\alpha ,Y} \sum\limits_{v = 1}^m {\alpha _v
	tr\left( {YL^{\left( v \right)}Y^T} \right)} \\
s.t.\;YY^T = I;\;\sum\limits_{v = 1}^m {\alpha _v } = 1,\;\;\alpha _v \ge 0
\\
\end{array}
\end{equation}

Where $L^{(v)}$ is an unnormalized graph Laplacian matrix for the $v$th view. It is the reflection of the neighborhood relationships between features in the $v$th views. $\alpha=[\alpha _1,\alpha _2,\cdot\cdot\cdot,\alpha _m]\in \Re^{m}$ is a set of nonnegative weights and $\alpha_v$ can reflect the importance of the $v$th view played for MSE to obtain the low-dimensional representations. $tr(\cdot)$ represents the trace of the matrix. MSE has verified that multi-view features contains more discriminative information and effectively explores the complementary property of different views to obtain an effective low-dimensional embedding for multi-view data sets. Meanwhile, authors argued that MSE achieve better performances facing with multimedia data but text data whose feature from different views are nearly independent.

\section{Co-regularized Multi-view Sparse Reconstruction Embedding}

In this section, we described our proposed CMSRE which finds a low-dimensional embedding over all views simultaneously. Given multi-view features with $n$ samples having $m$ views, $X=\{X^{(v)}\in \Re^{m_{v}\times n}\}^m_{v=1}$, where $X^{(v)} = [x^{(v)}_1, x^{(v)}_2,\cdot \cdot \cdot, x^{(v)}_n]\in \Re^{m_{v}\times n}$ is a feature matrix which consists of features from the $v$th view. $m_v$ is the dimension of features from the $v$th view. CMSRE aims to fully exploit the correlations of sparse correlations for all views and utilizes a co-regularized scheme to construct low-dimensional representations as $Y=\{Y^{(v)}\in \Re^{d\times n}\}^m_{v=1}$, where $Y^{(v)} = [y^{(v)}_1, y^{(v)}_2,\cdot \cdot \cdot, y^{(v)}_n]\in \Re^{d\times n}$ is the low-dimensional representations for the $v$th view and $d$ is dimension of common subspace constructed by CMSRE. 

\begin{figure*}[htbp]
	\centering
	\includegraphics[width=6in]{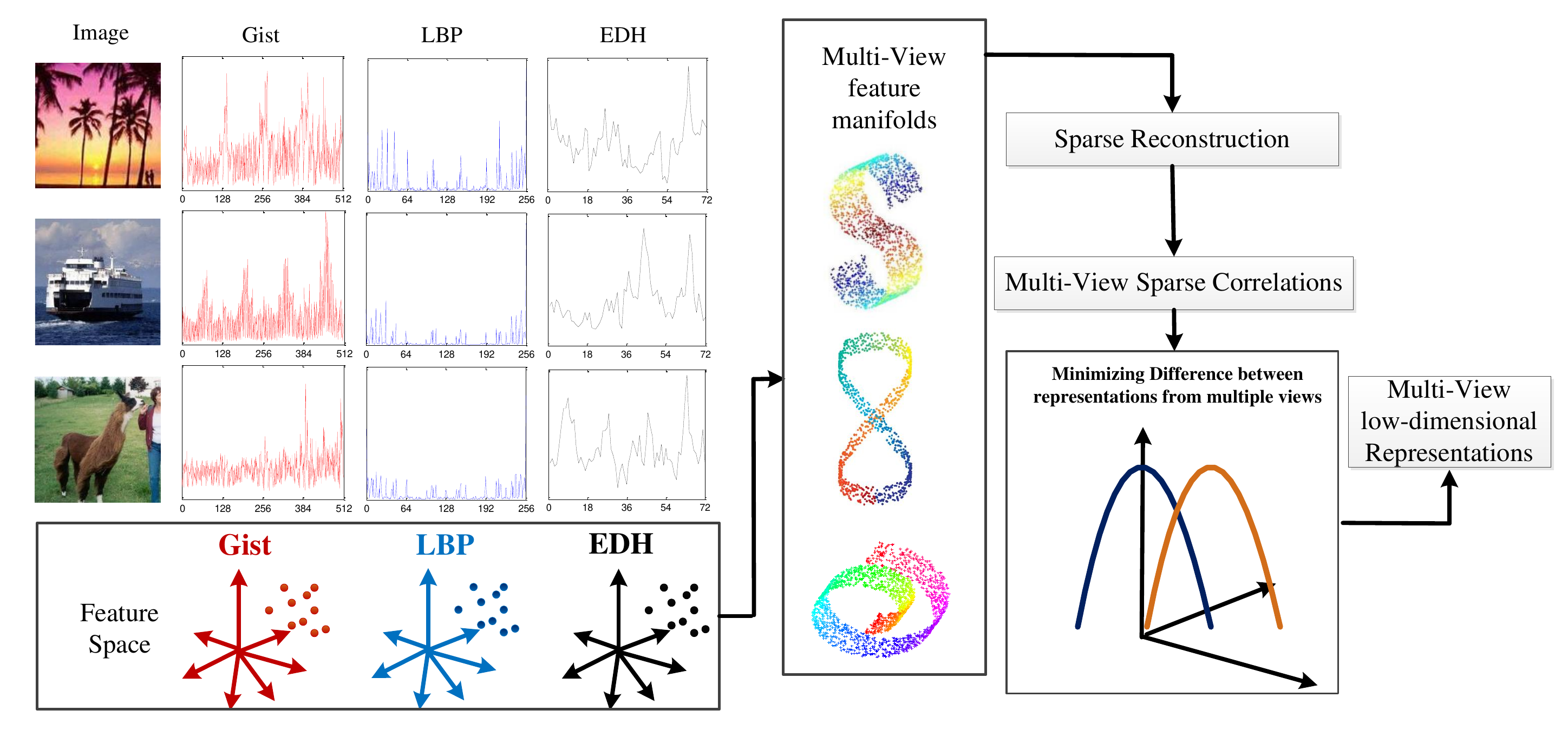}   
	\caption{The working procedure of CMSRE}
	\label{fig1}
\end{figure*}

Based on the aforementioned statements, CMSRE first builds a patch for each sample on a view. Then, based on the patches from different views, the single-view optimization is performed for CMSRE to get the optimal low-dimension representations for each view. Afterward, a co-regularized scheme is adopted by CMSRE to unify multiple views and fully exploits information from multi-view features. Finally, the solution of CMSRE is obtained by using an iterative optimization procedure. And Fig.\ref{fig1} shows the working procedure of CMSRE.

\subsection{Single-view Optimization}

In this section, we introduced how we conducted single-view optimization for CMSRE. Single-view optimization can help CMSRE to obtain best low-representations for features from each single view. Firstly, we utilized sparse representation~\cite{wright2009robust} to realize sparse reconstruction between each feature and its local neighbours as follows:

\begin{equation}
\label{eq2}
\begin{array}{l}
~~~~~\min\limits_{\tilde{s^{v}_i}}{\|\tilde{s^{v}_i}\|_1} \\
s.t. ~~ \textbf{1}^T {\tilde{s^{v}_i}}=1 \\
~~~~~\|x^{(v)}_i - X^{(v)}_{i}\tilde{s^{v}_i}\|_2 < \varepsilon
\end{array}
\end{equation}

Where $X^{(v)}_{i}\in \Re^{m_{v}\times k}$ is a subset of $X^{(v)}\in \Re^{m_{v}\times n}$ and contains the $k$ local neighbours of $x^{(v)}_i$ is the $i$th feature from the $v$th view. $\tilde{s^{v}_i}\in \Re^{d}$ reflects the correlations of sparse reconstruction between $x^{(v)}_i$ and its $k$ neighbours. It contains most information of its location and correlations with its neighbours. $\textbf{1}\in \Re^{d}$ is a vector of all ones and $\varepsilon$ is a small constant. The solution $\tilde{s^{v}_i}$ in Eq.\ref{eq2} is a sparse vector which can reconstruct each feature $x^{(v)}_i$ using its neighbours from the same view. 

For our proposed CMSRE, it's essential to fully exploit correlations of sparse reconstruction and maintain them into the low-dimensional representations. Therefore, CMSRE construct the following single-view optimization function which can maintain the correlations of sparse reconstruction for each view as much as possible: 

\begin{equation}
\label{eq3}
\begin{array}{l}
\mathop{\arg\min}\limits_{\{Y^{(v)}\}^{m}_{v=1}}\sum\limits_{v=1}^m \sum\limits_{i=1}^n \| y^{(v)}_i - Y^{(v)}_{i}\tilde{s^{v}_i}\|_2 \\
s.t.~~~~ {Y^{(v)}}^TY^{(v)} = \emph{\textbf{I}},~~ \forall v=1,2,\cdots,m 
\end{array}
\end{equation}

Where $Y^{(v)}_{i}$ is a subset of $Y^{(v)}$, which contains $k$ neighbours of $y^{(v)}_i$. $\emph{\textbf{I}}$ is a unit matrix whose diagonal elements are all ones, zeros the other. Eq.\ref{eq3} aims to find low-dimensional representations $\{Y^{(v)}\}^{m}_{v=1}$ for $\{X^{(v)}\}^{m}_{v=1}$. In order to obtain a more general expression, we rewrite Eq.\ref{eq3} as follows:

\begin{equation}
\label{eq4}
\begin{array}{l}
\mathop{\arg\min}\limits_{\{Y^{(v)}\}^{m}_{v=1}}\sum\limits_{v=1}^m \sum\limits_{i=1}^n \| y^{(v)}_i - Y^{(v)}s^v_i\|_2
\\
s.t.~~~~ {Y^{(v)}}^TY^{(v)} = \emph{\textbf{I}},~~ \forall v=1,2,\cdots,m 
\end{array}
\end{equation}

Where we utilized $Y^{(v)}s^v_i$ to replace $Y^{(v)}_{i}\tilde{s^{v}_i}$. The elements in $\tilde{s^{v}_i}\in \Re^k$ is contained in $s^v_i\in \Re^n$ at the corresponding locations while the other elements are supplemented using zeros. In order to obtain a more compact expression, CMSRE can be reorganized as follows and the inference process can be found in Appendix \uppercase\expandafter{\romannumeral1}.

\begin{equation}
\label{eq5}
\begin{array}{l}
\mathop{\arg\min}\limits_{\{Y^{(v)}\}^{m}_{v=1}}\sum\limits_{v=1}^m tr\left(Y^{(v)}M^{(v)}{(Y^{(v)})}^T\right)
\\
s.t.~~~~ {Y^{(v)}}^TY^{(v)} = \emph{\textbf{I}},~~ \forall v=1,2,\cdots,m 
\end{array}
\end{equation}

Where $M^{(v)} = (I-S^{(v)})(I-S^{(v)})^T \in \Re^{n\times n}$ and $S^{(v)}=[s^v_1,s^v_2,\cdots,s^v_n]\in \Re^{n\times n}$. $Y^{(v)}$ is the low-dimensional representations for features from the $v$th view. Because the smallest eigenvalue is close to $0$, $Y^{(v)}$ is formed by eigenvectors corresponding the smallest $2$ to $d+1$ eigenvalues. Because Eq.\ref{eq5} can calculate all optimal $Y^{(v)}$ from each single view without multi-view coordination, it's essential for us to develop a co-regularized scheme to integrate compatible and complementary information from multiple views.  

\subsection{Co-regularized Scheme}

In section \uppercase\expandafter{\romannumeral3}-$\it{A}$, we have shown how CMSRE conduct single-view optimization to obtain low-dimensional representations. However, single-view optimization only considers features from each single view and cannot integrate compatible and complementary information from multiple views. Even though low-dimensional representations preserve correlations of sparse reconstructions from each view, CMSRE fails to consider multi-view features from multiple views simultaneously. In this section, we illustrate the co-regularized scheme which is adopted by CMSRE to achieve multi-view learning. The co-regularized scheme provide a tool which can combine multiple kernels (or similarity matrices) from multiple views for multi-view dimension reduction problem. And we clearly show how we construct the co-regularized scheme for CMSRE in the following content. 

For CMSRE, the eigenvector matrix $Y^{(v)}$ contains low-dimensional representations for $v$th view. In the co-regularized scheme, we encourage the pairwise similarities of low-dimensional representations (in terms of rows of $Y^{(v)}$'s, $v=1,2,\cdots,m$) be similar across all the views.

Firstly, we work with two-view case for the ease of explanation and then extend it to multi-view situation. Because features from multiple views are all descriptors of one same sample, low-dimensional representations from different views should be in line with  consensus as much as possible. Therefore, we define the following cost function as a measure of disagreement between low-dimensional representations from two views:

\begin{equation}
\label{eq6}
D(Y^{(v)},Y^{(u)}) = \left\|\frac{K_{Y^{(v)}}}{\|K_{Y^{(v)}}\|^2_F}-\frac{K_{Y^{(u)}}}{\|K_{Y^{(u)}}\|^2_F} \right\|^2_F
\end{equation}

Where $K_{Y^{(v)}}$ is the similarity matrix for $Y^{(v)}$ and CMSRE defines the $i$th row with the $j$th column element of $K_{Y^{(v)}}$ is $k(y^{(v)}_i,y^{(v)}_j) =(y^{(v)}_i)^Ty^{(v)}_j$, which reflects the correlations between the $i$th and $j$th features from the $v$th view. Therefore, $K_{Y^{(v)}}=(Y^{(v)})^TY^{(v)}$. $\|\cdot\|_F$ is the Frobenius norms to make $\{K_{Y^{(v)}}, v=1,\cdots,m\}$  comparable across all views. In order to obtain a compact expression, Eq.\ref{eq6} can be further transfered as follows without considering the constant additive and scaling terms. And the inference process can be found in Appendix \uppercase\expandafter{\romannumeral2} :

\begin{equation}
\label{eq7}
D(Y^{(v)},Y^{(u)}) = -tr((Y^{(v)})^TY^{(v)}(Y^{(u)})^TY^{(u)})
\end{equation}

Because features from different views are all descriptors of one same sample, minimizing the disagreement $D(Y^{(v)},Y^{(u)})$ aims to integrate compatible and complementary information from the $v$th and $u$ views. And Due to Appendix \uppercase\expandafter{\romannumeral2}, it's equivalent to minimize $-tr((Y^{(v)})^TY^{(v)}(Y^{(u)})^TY^{(u)})$ to finish this goal. In order to combine it with single-view optimization, a co-regularized scheme is constructed and we get the following joint
minimization problem for two views:

\begin{equation}
\label{eq8}
\begin{array}{l}
\mathop{\arg\min}\limits_{Y^{(v)},{Y^{(u)}}} tr\left(Y^{(v)}M^{(v)}{(Y^{(v)})}^T\right) + tr\left(Y^{(u)}M^{(u)}{(Y^{(u)})}^T\right)\\
~~~~~~~~~ ~~~~~~~+\lambda D(Y^{(v)},Y^{(u)})
\\
s.t.~~~~ {Y^{(v)}}^TY^{(v)} = \emph{\textbf{I}},~~  {Y^{(u)}}^TY^{(u)} = \emph{\textbf{I}}
\end{array}
\end{equation}

Where $\lambda>0$ is a regularized parameter which balances the trade-off. Eq.\ref{eq8} simultaneously maintains sparse reconstructions from two views and minimizes disagreement between these two views. It can help features from two views to learn correlations of sparse reconstructions from each other to finish the cooperation for dimension reduction. And we always set $\lambda=0.8$ in our experiments. Then, in order to extend Eq.\ref{eq8} to multiple views, we further give the optimization objective as follows:

\begin{equation}
\small
\label{eq9}
\begin{array}{l}
\mathop{\arg\min}\limits_{\{Y^{(v)}\}^{m}_{v=1}}\sum\limits_{v=1}^m tr\left(Y^{(v)}M^{(v)}{(Y^{(v)})}^T\right) +  \lambda \sum\limits_{\substack{ 1 \leq v,u\leq m \\ v \neq u }}D(Y^{(v)},Y^{(u)}) \\
s.t.~~~~ {Y^{(v)}}^TY^{(v)} = \emph{\textbf{I}},~~ \forall v=1,2,\cdots,m
\end{array}
\end{equation}

$Y^{(v)}, v=1,2,\cdots,m$ can be calculated using the optimization problem as Eq.\ref{eq9}. And for each $Y^{(v)}$ from the $v$th view, it is constructed by considering the disagreements between the $v$th view and the other $m-1$ views. Through the co-regularized scheme, low-dimensional representations from all views have fully utilized information from the other views, which leads CMSRE to be meaningful for multi-view features. Furthermore, we illustrate the optimization procedure of CMSRE in the next part.

\subsection{Optimization Procedure of CMSRE}

In this section, we introduce the optimization procedure of CMSRE in detail. It is clearly that CMSRE aims to obtain the low-dimensional representations for features from multiple views simultaneously. The optimization procedure should optimize the solutions of CMSRE at the same time. Firstly, we utilize Eq.\ref{eq5} to initialize  $Y^{(v)}, v=1,2,\cdots,m$ which are trained without any information from other views. Then, in order to utilized Eq.\ref{eq9} to obtain $Y^{(v)}$ for the $v$th view, CMSRE maintain the low-dimensional representations from the other $m-1$ views to update $Y^{(v)}$ individually. By replace $D(Y^{(v)},Y^{(u)})$ using Eq.\ref{eq7}, CMSRE update $Y^{(v)}$ as follows:

\begin{equation}
\small
\label{eq10}
\begin{array}{l}
\mathop{\arg\min}\limits_{\{Y^{(v)}\}^{m}_{v=1}}\sum\limits_{v=1}^m tr\left(Y^{(v)}\left(M^{(v)}-\lambda \sum\limits_{\substack{ 1 \leq u\leq m \\ u \neq v }}  (Y^{(u)})^TY^{(u)} \right){(Y^{(v)})}^T\right) \\
s.t.~~~~ {Y^{(v)}}^TY^{(v)} = \emph{\textbf{I}},~~ \forall v=1,2,\cdots,m
\end{array}
\end{equation}

$Y^{(v)}$ can be calculated using eigen decomposition once all the other low-dimensional representations are determined. The optimization procedure utilizes iterative eigen decomposition to update low-dimensional representations from different views separately to obtain the optimal solution for all views. In order to help readers to understand the solving procedure, we show the optimization procedure as \ref{tab1}.

\begin{table}[htbp]
	\caption{The optimization procedure of CMSRE}
	\begin{center}
		\begin{tabular}{p{400pt}}
			\hline
			{Solving $Y=\{Y^{(v)}\in \Re^{m_{v}\times n}\}^m_{v=1}$ for all views} \\
			\hline
			\textbf{Input:}\\
			$X=\{X^{(v)}\in \Re^{m_{v}\times n}\}^m_{v=1}$: features from all views; $\lambda$: the regularized parameter.\\
			\textbf{Output:}\\
			$Y=\{Y^{(v)}\in \Re^{m_{v}\times n}\}^m_{v=1}$: the low-dimensional representations for all views.\\
			\textbf{Optimization Procedure:}\\
			1. Initialize all $Y=\{Y^{(v)}\in \Re^{m_{v}\times n}\}^m_{v=1}$ for all views using single-view optimization in Eq.\ref{eq5}.\\
			2. \textbf{Iterate}\\
			~~2.1. Utilized sparse representation to construct correlation between neighbors $\{S^{(v)}\in \Re^{n \times n}\}^m_{v=1}$ for all views.\\
			~~2.2. Construct $\{M^{(v)}\in \Re^{n \times n}\}^m_{v=1}$ using $S^{(v)}$ for all views.\\
			~~2.3. Combining $\{M^{(v)}\in \Re^{n \times n}\}^m_{v=1}$ with $D(Y^{(v)},Y^{(u)})$ in Eq.\ref{eq7}, $Y^{(v)}$ can be calculated according to Eq.\ref{eq10}.\\
			
			3. \textbf{Until Convergence}\\
			4. \textbf{Return} $Y=\{Y^{(v)}\in \Re^{m_{v}\times n}\}^m_{v=1}$\\
			
			\hline
		\end{tabular}
		\label{tab1}
	\end{center}
\end{table}

\section{Experiment}

In order to show the excellent performance of our proposed method, we conduct various experiments(including document classification, image retrieval, etc.) on several benchmark multi-view datasets in this paper. All these experiments can verify that our proposed CMSRE achieves better performances in most situations.

\subsection{Datasets and comparing methods}

\begin{figure*}[htbp]
	\centering
	\includegraphics[width=4in]{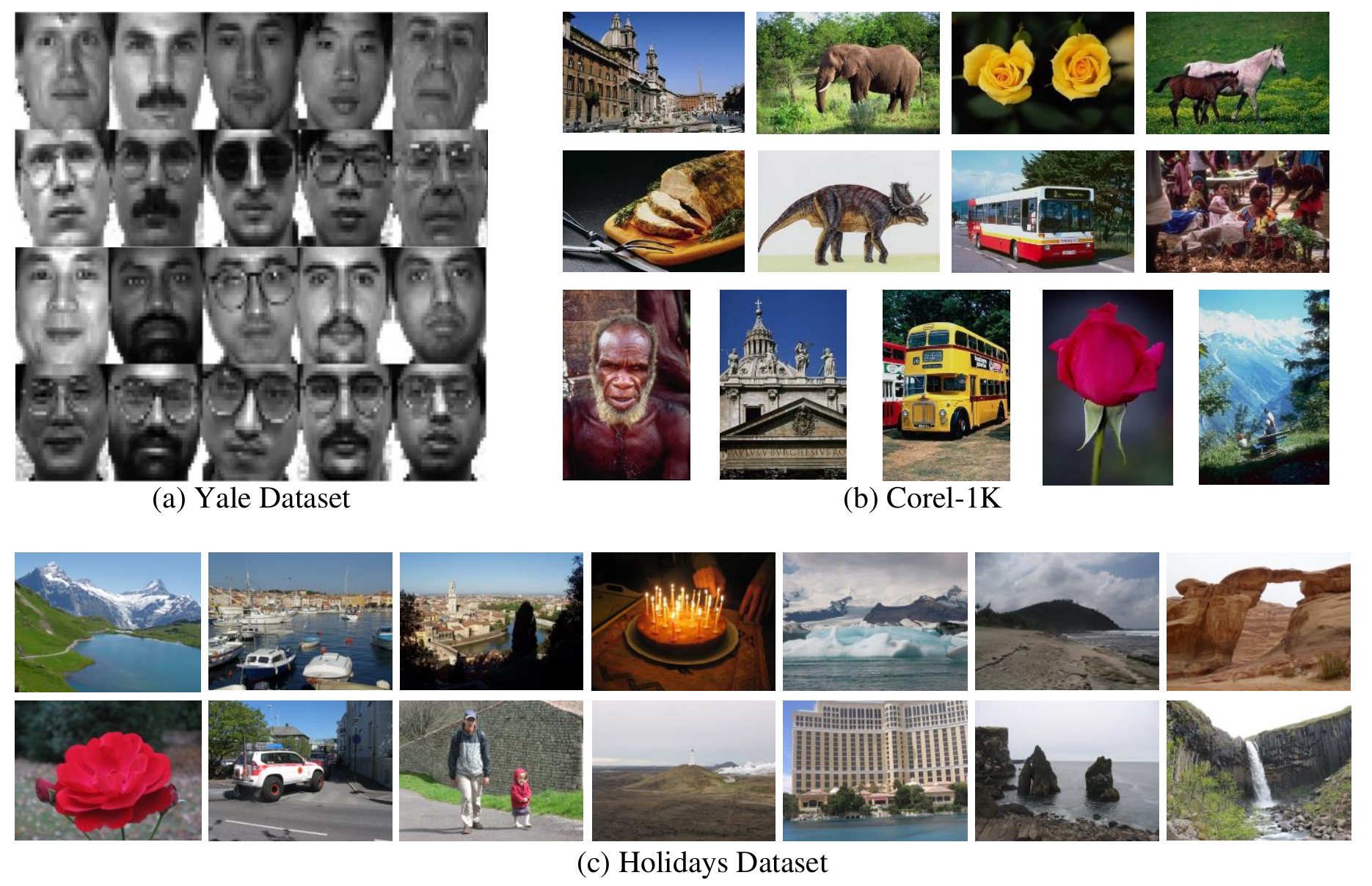}   
	\caption{Images from datasets utilized in our experiment}
	\label{fig2}
\end{figure*}

In our experiments, 6 datasets are utilized to show the excellent performances of CMSRE, including document datasets (such as, 3Sources\footnote{http://mlg.ucd.ie/datasets/3sources.html}, Cora\footnote{http://lig-membres.imag.fr/grimal/data.html}) and face datasets (Yale\footnote{http://cvc.yale.edu/projects/yalefaces/yalefaces.html}), and image datasets (such as, Holidays\footnote{http://lear.inrialpes.fr/~jegou/data.php}\cite{jegou2008hamming} and Corel-1K\footnote{https://sites.google.com/site/dctresearch/Home/content-based-image-retrieval}). All document datasets are benchmark multi-view datasets. For those image datasets, we extracted features using multiple descriptors as multi-view features for our experiments. Some images from these image datasets are shown as Fig.\ref{fig2}.


The performance of CMSRE is evaluated by comparing the following methods: 1. Pairwise\cite{kumar2011coreg}, which is a multi-view spectral embedding methods using pairwise constraints. 2. Centroid\cite{kumar2011coreg}, which utilizes centroid constraints for multi-view learning. 3. MSE\cite{xia2010multiview}. 4. LE\_BSV which is laplacian eigenmaps\cite{belkin2003laplacian} with best view. 5. LLE\_BSV is locally linear embedding\cite{roweis2000nonlinear} with best view. 6. PCA\_BSV is principle component analysis with best view.

\subsection{Document classification}

In this section, in order to show the advantages CMSRE on document classification, we conducted 2 experiments on benchmark multi-view document datasets, including 3Sources and Cora. 

3Sources was collected from 3 online news sources, BBC, Reuters and Guardian. Each source was treated as one view in this dataset. All samples have three views and are utilized to train all multi-view DR methods in our experiments. Then, among all 169 samples, 69 samples are randomly assigned as those ones which need to be classified. After all DR methods, 1NN classifier is adopted to show the classification results. And we conducted this experiment for 20 times to show the mean and max results as Table.\ref{tab3}:

\begin{table}[htbp]
	\center
	\caption{Classification Accuracies (\%) on 3Sources}
	\begin{tabular}{ccccccc}
		\hline
		\raisebox{-1.50ex}[0cm][0cm]{Methods}&
		\multicolumn{2}{c}{Dimension=10} &
		\multicolumn{2}{c}{Dimension=20} &
		\multicolumn{2}{c}{Dimension=30}  \\
		\cline{2-7}
		&
		Mean&
		Max&
		Mean&
		Max&
		Mean&
		Max \\
		
		\hline
		CMSRE& \textbf{84.58} & \textbf{89.86} &  \textbf{84.64}& \textbf{92.75}& \textbf{86.16} & \textbf{94.20}\\
		Pariwise\cite{kumar2011coreg}& 79.36 &85.51& 80.62& 89.86& 82.61& 91.30 \\
		Centroid\cite{kumar2011coreg}& 82.00& 86.96& 82.90& 89.86&  85.29& 89.86\\
		MSE&79.14 & 85.51 & 79.29& 88.41& 84.57& 89.86\\ 
		LE\_BSV& 81.59& 88.41 & 82.46 & 89.86& 83.86& 86.96\\ 
		LLE\_BSV& 78.04& 85.51 & 80.66& 86.96& 82.17& 88.41 \\ 
		PCA\_BSV& 80.02& 88.41&   80.81& 89.86& 84.06 & 89.86\\
		\hline
	\end{tabular}
	\label{tab3}
\end{table}

Cora consists of 2708 scientific publications which come from 7 classes. Each document is represented by content and cites views. Among all 2708 samples, 708 samples are randomly assigned as those ones which need to be classified. After all DR methods, 1NN classifier is adopted to show the classification results. And we conducted this experiment for 20 times to show the mean and max results as Table.\ref{tab4}:

\begin{table}[htbp]
	\center
	\caption{Classification Accuracies (\%) on Cora}
	\begin{tabular}{ccccccc}
		\hline
		\raisebox{-1.50ex}[0cm][0cm]{Methods}&
		\multicolumn{2}{c}{Dimension=10} &
		\multicolumn{2}{c}{Dimension=20} &
		\multicolumn{2}{c}{Dimension=30}  \\
		\cline{2-7}
		&
		Mean&
		Max&
		Mean&
		Max&
		Mean&
		Max \\
		
		\hline
		CMSRE& \textbf{55.79} & \textbf{60.03} &  \textbf{66.46}& \textbf{70.90}& \textbf{68.35} & \textbf{71.61}\\
		Pariwise\cite{kumar2011coreg}& 50.06 & 54.52& 51.73  & 55.93 & 52.24  & 56.15  \\
		Centroid\cite{kumar2011coreg}& 50.82 & 55.79 & 56.49  & 60.17  &  58.56 & 63.42 \\
		MSE&  53.10 &  57.49  & 58.36  & 62.85  & 58.99  &  62.99 \\ 
		LE\_BSV&  44.34 & 47.74  &  57.05  &  61.02 &  57.48 &  61.30 \\ 
		LLE\_BSV& 46.67  &  50.71  & 50.67  & 55.79  & 51.38  &  55.93 \\ 
		PCA\_BSV& 51.61 & 55.23  & 52.41 & 56.92  &  52.87  & 58.19  \\
		\hline
	\end{tabular}
	\label{tab4}
\end{table}

It can be found easily that our proposed CMSRE can achieve best performances to deal with these two document datasets. Meanwhile, experiment results show that performances of single-view methods are always lower than those multi-view learning ones, which can verify that multi-view learning is a valuable topic and need further researches.

\subsection{Face Recognition}

In this section, we carried on a face recognition experiment on Yale dataset to show the effectiveness of our method. There are 165 faces corresponding to 15 peoples in Yale datasets. We extract features using gray-scale intensity, local binary patterns \cite{ojala2002multiresolution} and edge direction histogram \cite{gao2008image} as 3 views. Among all 165 faces, we randomly selected 65 faces as those ones to be recognized. After all DR methods, 1NN classifier is adopted to calculate the recognition results. This experiment has been conducted for 20 times and we show the mean and max recognition accuracies as Table.\ref{tab5}.

\begin{table}[htbp]
	\center
	\caption{Recognition Accuracies (\%) on Yale Faces}
	\begin{tabular}{ccccccc}
		\hline
		\raisebox{-1.50ex}[0cm][0cm]{Methods}&
		\multicolumn{2}{c}{Dimension=10} &
		\multicolumn{2}{c}{Dimension=20} &
		\multicolumn{2}{c}{Dimension=30}  \\
		\cline{2-7}
		&
		Mean&
		Max&
		Mean&
		Max&
		Mean&
		Max \\
		
		\hline
		CMSRE& \textbf{70.38} & \textbf{80.00 } &  75.08& 87.69& \textbf{ 80.78 } & \textbf{ 90.77 }\\
		Pariwise\cite{kumar2011coreg}& 62.46  & 72.31 & 72.68   & 86.15  & 73.98   &  87.69  \\
		Centroid\cite{kumar2011coreg}&  62.92 & 69.23  & \textbf{76.53}  & \textbf{88.17}   &  77.78  & 90.77  \\
		MSE& 59.69  &  66.15  &  65.90  &76.92   & 66.99  & 83.08\\ 
		LE\_BSV&  53.92  &  63.08  &  60.60   &  76.92  &  58.78  &  73.85  \\ 
		LLE\_BSV&  48.08  &  56.92   & 57.42   &  72.31  &  57.88  & 72.31   \\ 
		PCA\_BSV&  56.69 &  63.08  & 60.58  & 73.01   &  60.65  &  73.85 \\
		\hline
	\end{tabular}
	\label{tab5}
\end{table}

This experiment can verify that CMSRE is a good multi-view dimension reduction method for face recognition. It can achieve good performances in most situations. Meanwhile, Centroid\cite{kumar2011coreg} is also a good method to deal with multi-view datasets. 

\subsection{Image Retrieval}

In this section, we conducted  2 experiments on different image datasets (including Holidays and Corel-1k datasets) for image retrieval, which can verify the excellent performances of CMSRE. 

For Holidays dataset, there are 1491 images corresponding to 500 categories, which are mainly captured for sceneries. Among all these images, 500 images are assigned as the query images while the other 991 are assigned as the corresponding relevant images. We utilized MSD~\cite{liu2011image}, Gist~\cite{oliva2001modeling} and HOC~\cite{yu2016novel} to extract features as 3 views for all images. And all these methods were conducted to project all samples to a 50-dimensional subspace. We selected the best view for all these methods. Distance metric~\cite{wang2016semantic} is essential for image retrieval and  utilized $L1$ distance to measure similarities between samples. This experiment has been conducted 20 times and we show the mean results of precision, recall, mean average precision (MAP) and $F_1$-Measure in Table \ref{tab6}.

\begin{table*}[htbp]
	\center
	\caption{The precision ($P\%$), recall ($R\%$), MAP ($\%$) and $F_1$-Measure of different methods on Holidays dataset.}
	\begin{tabular}
		{cccccc}
		\hline
		\diagbox[width=5.8em]{\tiny{Methods}}{\tiny{Criteria}}
		&
		$P$&$R$&MAP& $F_1$\\
		\hline
		CMSRE & \textbf{79.44} & \textbf{61.37} & \textbf{89.72}& \textbf{35.56}\\
		
		Pairwise\cite{kumar2011coreg} &78.04 & 60.06& 88.92& 33.94 \\
		
		Centroid\cite{kumar2011coreg} &77.14 & 59.39& 88.47& 33.56\\
		
		MSE & 77.25 & 59.65 & 88.63& 33.66 \\
		
		LE\_BSV & 72.91 & 56.15& 86.42& 31.72 \\
		
		LLE\_BSV & 73.52 & 58.12& 87.56& 32.46 \\
		
		PCA\_BSV & 71.65 & 55.15& 82.25& 31.36 \\
		\hline
	\end{tabular}
	\label{tab6}
\end{table*}

\begin{figure*}[htp]
	\centerline{
		\subfloat[Precision]{\includegraphics[width=3in]{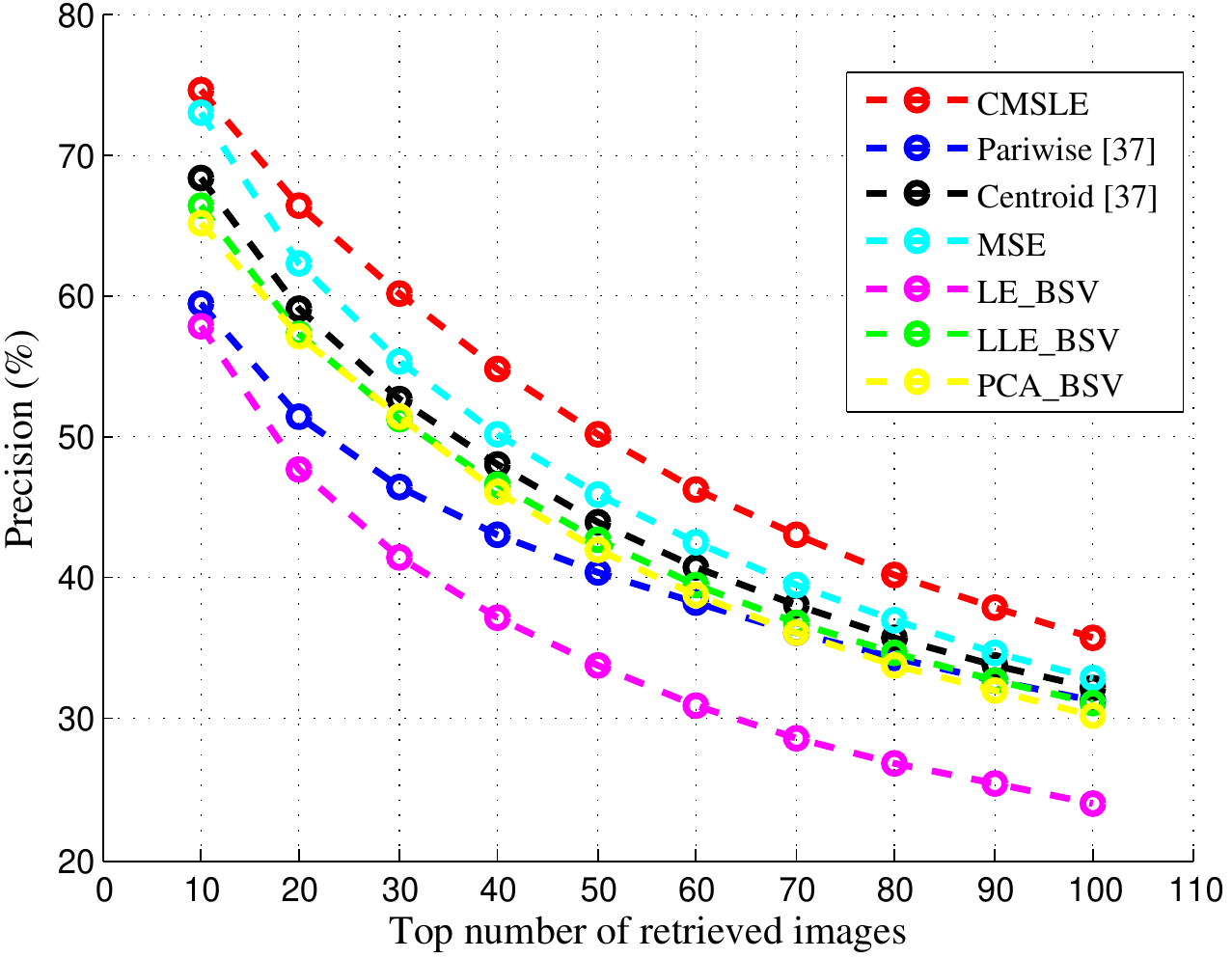}}
		\subfloat[Recall]{\includegraphics[width=3in]{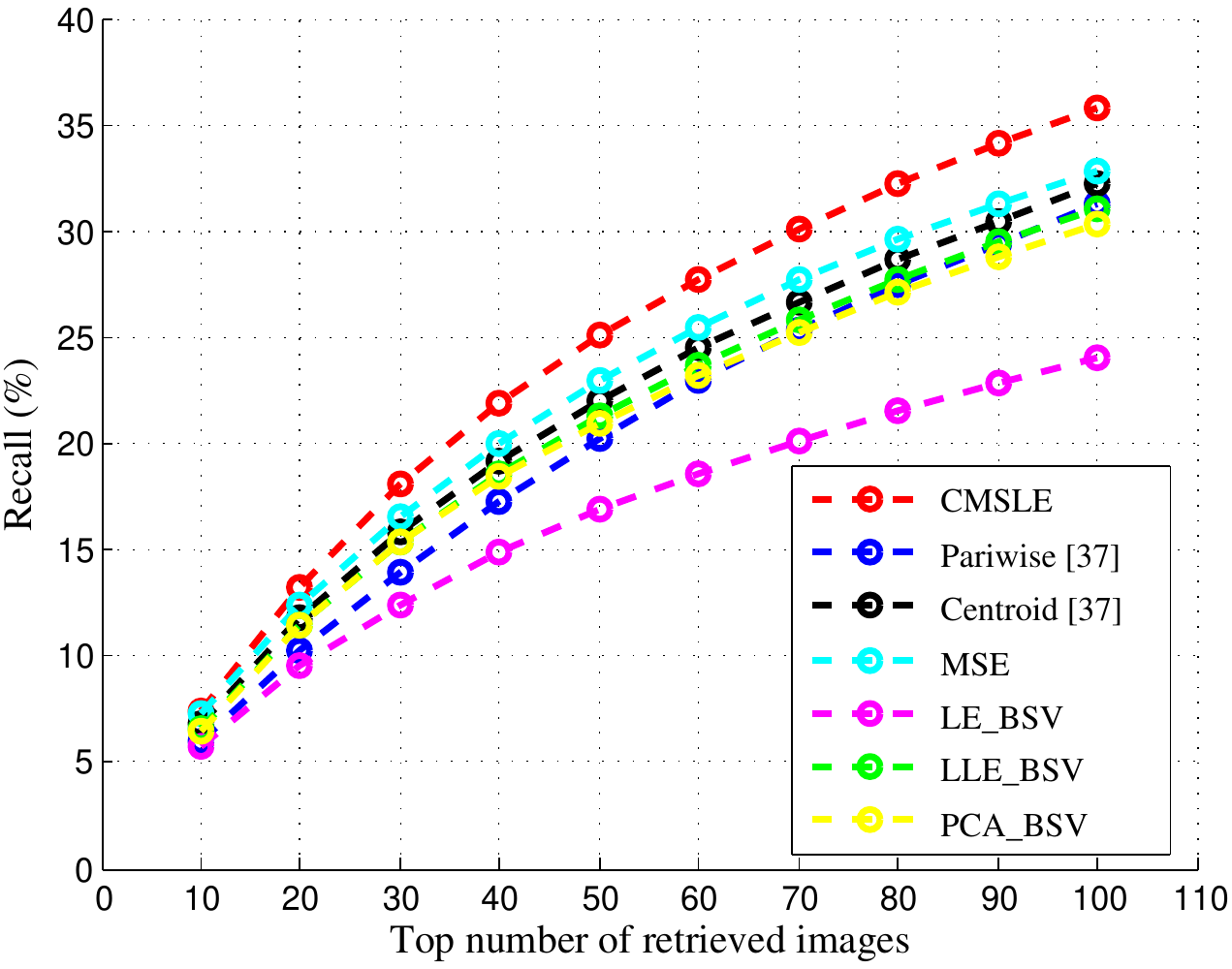}}}
	\centerline{
		\subfloat[PR-Curve]{\includegraphics[width=3in]{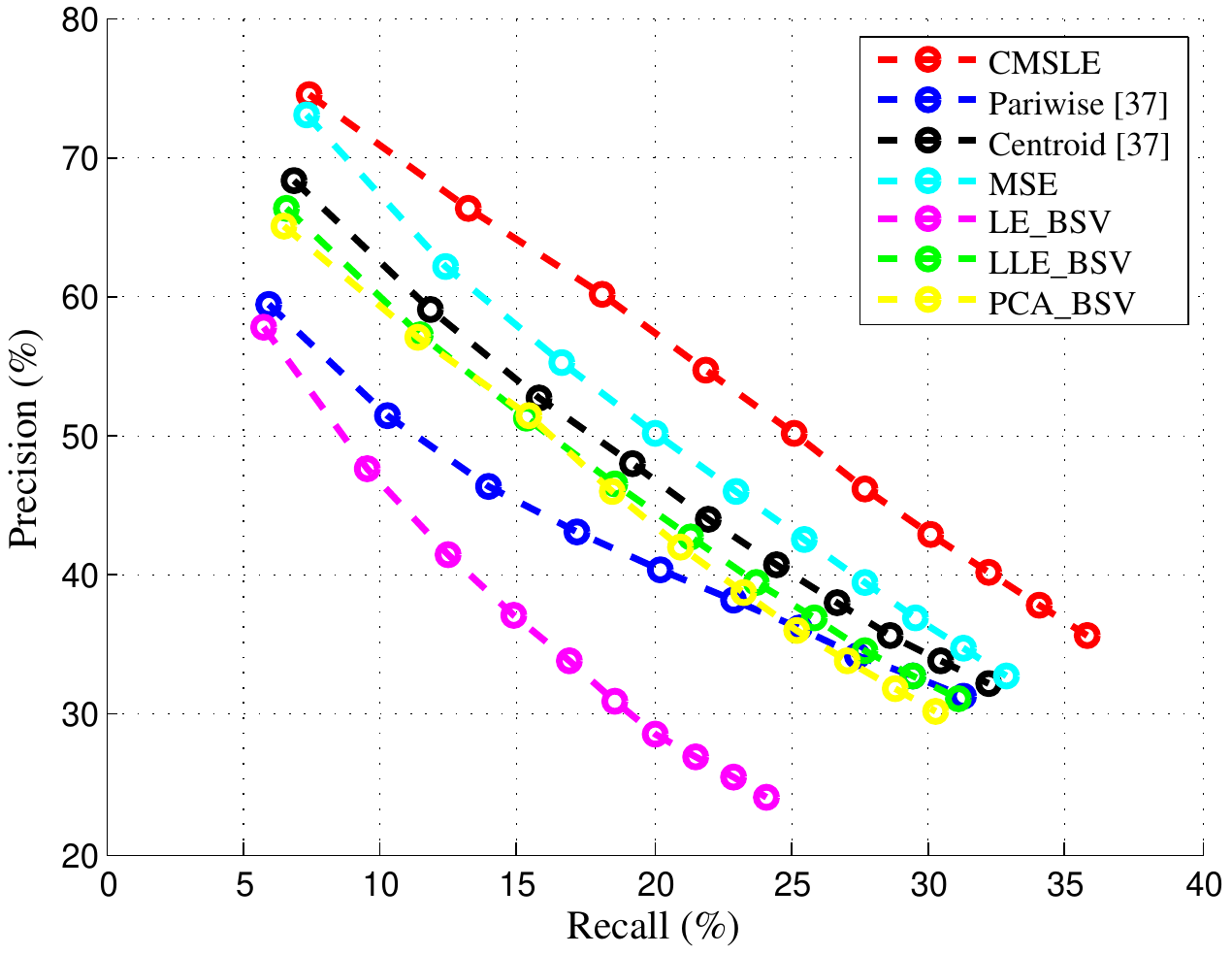}}
		\subfloat[$F_1$-Measure]{\includegraphics[width=3in]{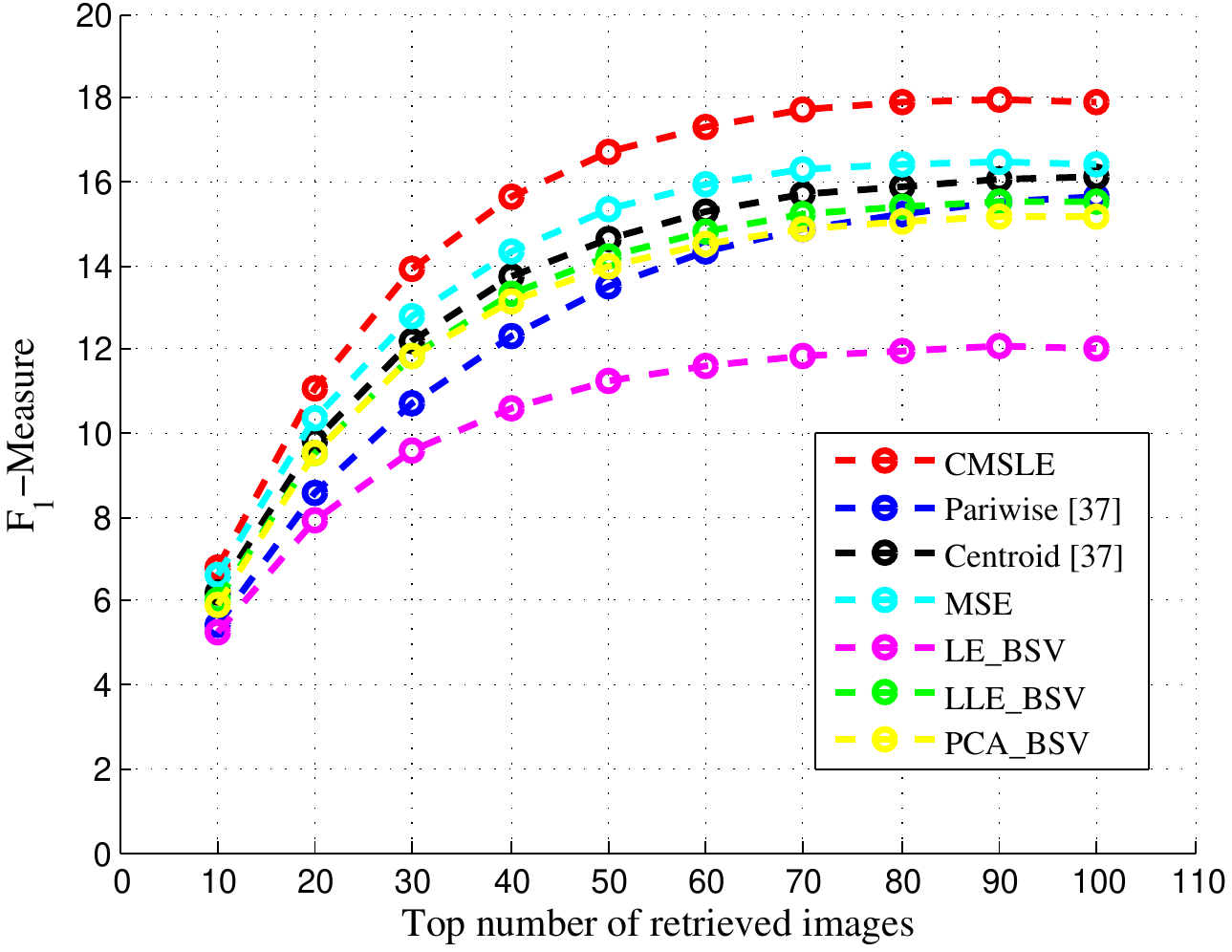}}
	}
	\caption{The curves of precision, recall, PR, and $F_1$-Measure on Corel-1k dataset. (This figure is best viewed in color)}
	\label{fig3}
\end{figure*}

It can be found that our proposed CMSRE is the best method to deal with multi-view features. Meanwhile, Pariwise \cite{kumar2011coreg} can also achieve good performance in most situations. And those single-view methods methods always perform poorly.

For Corel-1k dataset, there are 1000 images corresponding to 10 categories, which are collected just for image retrieval. For each category, we randomly select 10 images as query ones. We utilized the 3 features descriptors \cite{liu2011image,oliva2001modeling,yu2016novel} to extract features. All the methods were conducted to project all samples to a 50-dimensional subspace. The procedure of this experiment is same with that one on Holidays. We conducted this experiment 20 times and show the results as Fig.\ref{fig3}.  

Through these 2 experiments for image retrieval, we can clearly find that our proposed CMSRE outperforms the other DR methods in most situations. CMSRE can integrate compatible and complementary information from multi-view features and obtain better subspaces from these views. And the first 4 multi-view methods outperform the other single-view methods, which can show multi-view learning is a valuable research field indeed.

\subsection{Convergence Analysis of CMSRE}

Because CMSRE utilized iterative optimization procedure to obtain its solution, we should show the convergence property of CMSRE in this section. Therefore, we designed an experiment on Yale dataset, which is according to TABLE \ref{tab5}. We still project features into a 30-dimensional subspace and show the objective values and recognition accuracies with the increase of iterations. It is obvious that the change variation trend of objective values can show the convergence properties of CMSRE as Fig.\ref{fig4}:

\begin{figure}[htbp]
	\centerline{
		\includegraphics[width=4in]{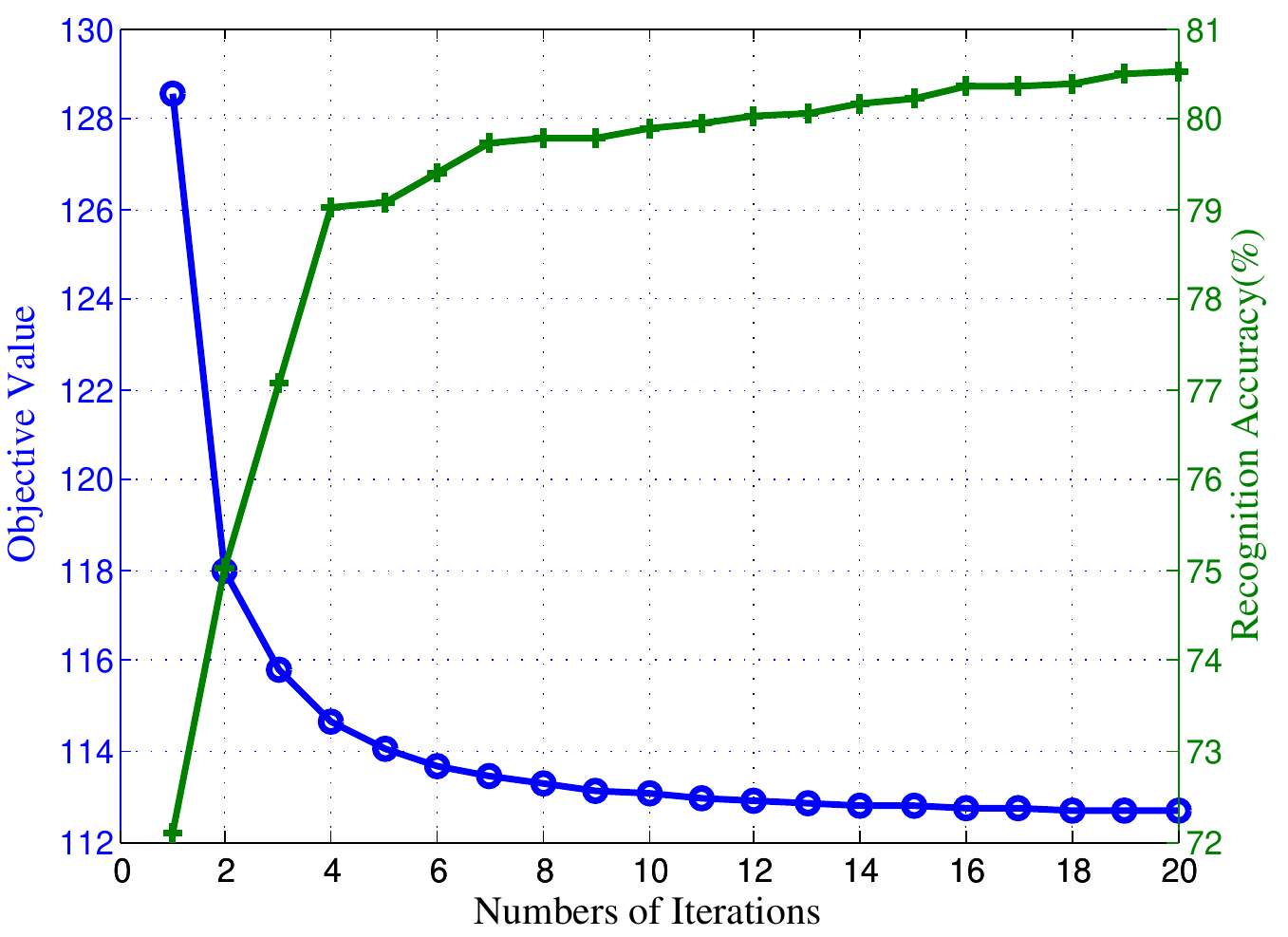}
	}
	\caption{The objective values and recognition accuracies with the number of iterations. (This figure is best viewed in color)}
	\label{fig4}
\end{figure}

It can be found that the objective values of CMSRE tend to be stable when 10 iterations finished. The recognition accuracies of Yale dataset is raising with the increase of iterations. Therefore, Fig.\ref{fig4} can verify that CMSRE converges once enough iterations are finished. 

\subsection{The Impact of the Regularized Parameter $\lambda$}

In order to show the impact of the regularized parameter $\lambda$, we conducted another experiment to show the performances of CMSRE with different values of $\lambda$. The experiment is based on the former experiment on Yale dataset. The experimental settings are the same as before in section 4.3. This experiment utilizes CMSRE with different $\lambda$ to project features into a 30-dimensional subspace. And we conducted this experiment for 20 times and showed the mean results as Table \ref{tab7}.

\begin{table}[htbp]
	\center
	\caption{Recognition Accuracies (\%) with different values of $\lambda$ for CMSRE}
	\begin{tabular}{ccccccc}		
		
		\hline
		Parameter& $\lambda = 0$ & $\lambda =0.5$ &  $\lambda =0.6$& $\lambda =0.7$& \textbf{$\lambda =0.8$} & $\lambda =0.9$\\
		
		\hline
		Accuracy & 65.85 & 74.31 &  76.78 & 78.32& \textbf{80.78} & 80.03\\
		\hline
	\end{tabular}
	\label{tab7}
\end{table}

It can be found clearly that value of $\lambda$ can influence the performance of CMSRE to some extent. And the performance of CMSRE is best when $\lambda = 0.8$. Therefore, we set $\lambda = 0.8$ in our experiment. Meanwhile, CMSRE degenerates into single view method when $\lambda=0$ and the performance of CMSRE is still better than LLE\_BSV. It is because CMSRE can exploit the correlations of sparse reconstruction for these views.

\section{Conclusion}

In this paper, we proposed a novel method named Co-regularized Multi-view Sparse Reconstruction Embedding (CMSRE) to deal with features from multiple views. By exploiting correlations of sparse reconstruction from multiple views, CMSRE is able to learn local sparse structures of nonlinear manifolds from multiple views and constructs significative low-dimensional representations for them. Furthermore, in order to preserve correlations of sparse reconstructions from multiple views, CMSRE proposes a co-regularized scheme to integrate compatible and complementary information from these features to construct low-dimensional subspace directly. Due to the novel construction of CMSRE, it can be utilized to deal with multi-view features for dimension reduction. Various experiments have shown that our proposed CMSRE can achieve excellent performances.

\section*{Acknowledge}

We would like to thank the anonymous reviewers for their valuable comments and suggestions to significantly improve the quality of this paper. This work was supported in part by the National Natural Science Foundation of China Grant 61370142 and Grant 61272368, by the Fundamental Research Funds for the Central Universities Grant 3132016352, by the Fundamental Research of Ministry of Transport of P. R. China Grant 2015329225300.

\section*{Appendix \uppercase\expandafter{\romannumeral1}}

This appendix shows how to obtain Eq.\ref{eq5} from Eq.\ref{eq4} using matrix operations.
\begin{equation}
\label{Appendix1}
\begin{array}{l}
~~~\sum\limits_{v=1}^m \sum\limits_{i=1}^n \| y^{(v)}_i - Y^{(v)}s^v_i\|_2
\\
=\sum\limits_{v=1}^m \sum\limits_{i=1}^n tr \left( (y^{(v)}_i-Y^{(v)}s^v_i)(y^{(v)}_i-Y^{(v)}s^v_i)^T\right)\\
=\sum\limits_{v=1}^m \sum\limits_{i=1}^n tr (y^{(v)}_i{y^{(v)}_i}^T-y^{(v)}_i{s^v_i}^T {Y^{(v)}}^T\\
~~~~~~~~~- Y^{(v)}s^v_i{y^{(v)}_i}^T+Y^{(v)}s^v_i{s^v_i}^T{Y^{(v)}}^T )\\
= \sum\limits_{v=1}^m tr(Y^{(v)}{Y^{(v)}}^T+Y^{(v)}S^{(v)}{S^{(v)}}^T{Y^{(v)}}^T \\
~~~~~~~~~- Y^{(v)}{S^{(v)}}^T{Y^{(v)}}^T - Y^{(v)}{S^{(v)}}{Y^{(v)}}^T)\\
=\sum\limits_{v=1}^m tr(Y^{(v)}(I-S^{(v)}(I-S^{(v)})^T {Y^{(v)}}^T)\\
=\sum\limits_{v=1}^m tr(Y^{(v)}M^{(v)}{Y^{(v)}}^T)
\end{array}
\end{equation}

Where $M^{(v)} = (I-S^{(v)})(I-S^{(v)})^T \in \Re^{n\times n}$ and $S^{(v)}=[s^v_1,s^v_2,\cdots,s^v_n]\in \Re^{n\times n}$. 

\section*{Acknowledge \uppercase\expandafter{\romannumeral2}}

Because $Y^{(v}), v=1,2,\cdots,m$ from different views are all orthogonal matrices, $\|K_{Y^{(v)}}\|^2_F = k, v=1,2,\cdots,m$. Therefore, we can transfer Eq.\ref{eq6} as follows:

\begin{equation}
\label{Appendix2}
\begin{array}{l}
D\left(Y^{(v)},Y^{(u)}\right) = \left\|\frac{K_{Y^{(v)}}}{\|K_{Y^{(v)}}\|^2_F}-\frac{K_{Y^{(u)}}}{\|K_{Y^{(u)}}\|^2_F} \right\|^2_F\\
=\frac{1}{k^2}\left\|K_{Y^{(v)}}-K_{Y^{(u)}}\right\|^2_F\\
=\frac{1}{k^2}\left\|(Y^{(v)})^TY^{(v)} - (Y^{(u)})^TY^{(u)} \right\|^2_F\\
=\frac{1}{k^2}tr\left(2kI-2(Y^{(v)})^TY^{(v)}(Y^{(u)})^TY^{(u)}\right)\\
=\frac{2}{k}-\frac{2}{k^2}tr\left((Y^{(v)})^TY^{(v)}(Y^{(u)})^TY^{(u)}\right)\\
\end{array}
\end{equation}

Where $I$ is a unit matrix whose diagonal elements are all $1$, $0$ others. By ignoring the constant additive and scaling terms, we can utilize $-tr((Y^{(v)})^TY^{(v)}(Y^{(u)})^TY^{(u)}$ to represent $D(Y^{(v)},Y^{(u)})$ and minimize it to shrink the disagreement between low-dimensional representations between the $v$th view and the $u$th view.



\section*{Reference}
\bibliographystyle{elsarticle-num}
\bibliography{Reference}


%
%
%
\end{document}